\documentclass[conference,letterpaper]{IEEEtran}
\IEEEoverridecommandlockouts
\usepackage{cite}
\usepackage{amsmath,amssymb,amsfonts}
\usepackage{algorithmic}
\usepackage{graphicx}
\usepackage{textcomp}
\usepackage{xcolor}
\def\BibTeX{{\rm B\kern-.05em{\sc i\kern-.025em b}\kern-.08em
  T\kern-.1667em\lower.7ex\hbox{E}\kern-.125emX}}
  \usepackage{amssymb}
\usepackage{bigstrut}
\usepackage{multirow}
\usepackage{graphicx}
\usepackage{algorithm}
\ifCLASSOPTIONcompsoc
 \usepackage[caption=false,font=normalsize,labelfont=sf,textfont=sf]{subfig}
\else
 \usepackage[caption=false,font=footnotesize]{subfig}
\fi

\begin{document}

\title{Large Scale Global Optimization Algorithms for IoT Networks: A Comparative Study\\
}

\author{\IEEEauthorblockN{Sotirios K. Goudos\IEEEauthorrefmark{1}, Achilles D. Boursianis\IEEEauthorrefmark{1},
Ali Wagdy Mohamed\IEEEauthorrefmark{2}, Shaohua Wan\IEEEauthorrefmark{3}, Panagiotis Sarigiannidis\IEEEauthorrefmark{4}, \\George K. Karagiannidis\IEEEauthorrefmark{5}, and Ponnuthurai N. Suganthan\IEEEauthorrefmark{6}}
\IEEEauthorblockA{\IEEEauthorrefmark{1}
ELEDIA@AUTH,School of Physics,Aristotle University of Thessaloniki,Thessaloniki, Greece}
\IEEEauthorblockA{\IEEEauthorrefmark{2}Operations Research Department, Faculty of Graduate Studies for Statistical Research,Cairo University, Giza,  Egypt,\\
School of Engineering and Applied Sciences, Wireless Intelligent Networks Center (WINC),\\ Nile University, Giza, Egypt}
\IEEEauthorblockA{\IEEEauthorrefmark{3}School of Information and Safety Engineering, Zhongnan University of Economics and Law, Wuhan, China}
\IEEEauthorblockA{\IEEEauthorrefmark{4}Department of Informatics and Telecommunications Engineering, University of Western Macedonia, Kozani, Greece}
\IEEEauthorblockA{\IEEEauthorrefmark{5}School of Electrical and Computer Engineering, Aristotle University of Thessaloniki,Thessaloniki, Greece}
\IEEEauthorblockA{\IEEEauthorrefmark{6}School of Electrical Electronic Engineering,Nanyang Technological University, Singapore}

}

\maketitle

\begin{abstract}
The advent of Internet of Things (IoT) has bring a new era in communication technology by expanding the current inter-networking services and enabling the machine-to-machine communication. IoT massive deployments will create the problem of optimal power allocation. The objective of the optimization problem is to obtain a feasible solution that minimizes the total power consumption of the WSN, when the error probability at the fusion center meets certain criteria. This work studies the optimization of a wireless sensor network (WNS) at higher dimensions by focusing to the power allocation of decentralized detection. More specifically, we apply and compare four algorithms designed to tackle Large scale global optimization (LGSO) problems. These are the memetic linear population size reduction and semi-parameter adaptation (MLSHADE-SPA), the contribution-based cooperative coevolution recursive differential grouping (CBCC-RDG3), the differential grouping with spectral clustering-differential evolution cooperative coevolution (DGSC-DECC), and the enhanced adaptive differential evolution (EADE). To the best of the authors knowledge, this is the first time that LGSO algorithms are applied to the optimal power allocation problem in IoT networks. We evaluate the algorithms performance in several different cases by applying them in cases with 300, 600 and 800 dimensions.
\end{abstract}


\section{Introduction}
Large scale global optimization (LGSO) problems are attracting significant attention by the researchers over the last years. Several optimization problems in wireless communications can be inherently extended to higher dimensions. The Internet of Things (IoT) paradigm, as well as the advent of 5G cellular communication systems, will use hundreds of wireless sensor nodes in a limited space that require an optimal power allocation scheme. These sets of wireless sensors work together to monitor an area, gather data from specific applications, and constitute a wireless sensor network (WSN). There is a huge number of applications that WSNs are utilized. Among the most representative examples are included the monitoring of environmental (both indoors/outdoors) parameters, the monitoring of the power grid, the tracking and targeting of military targets, the seismic waves sensing, and the healthcare and human activity monitoring \cite{Yick2008, Nadeem2015, Hsia2012}.

In wireless sensor networks, if the detection mechanism is based on a decentralized scheme, each sensor node, after a local pre-processing of its own observations, transmits a summary of the pre-processed data to the \emph{fusion center}, i.e. the central node of the wireless sensor architecture. Then, the central node applies a final decision based on a set of few hypotheses. Taking into consideration the given description of the decentralized scheme, the fusion center does not have direct access to the initial un-processed observation data. Thus, the data rate requirements of the decentralized scheme are significantly lower than in a centralized scheme (where all the nodes of the WSN send all the information to the fusion center directly). This is because only a few bits of the observation data are transmitted \cite{Tsitsiklis1993}. This aforementioned problem, i.e the distributed detection and fusion under specific constraints in WSNs, has attracted a lot of attention in the literature \cite{Appadwedula2005, Jayaweera2007, Krasnopeev2005}.

In this paper, we assume that the fusion center of a WSN is decentralized and operates on a binary hypothesis testing problem. Also, the observations of the wireless sensor nodes are correlated. In this context, we can define the goal of the optimization problem as the minimization of the total power, under the application of specific criteria to the error probability detection of the central node, having the optimal allocation of the total power resources.

The application of EAs to this problem has already been discussed in the literature by utilizing the Particle Swarm Optimization (PSO) \cite{Wimalajeewa2008}, the Differential Evolution (DE)\cite{WSNDE}, a hybrid Biogeography Based Optimization - Differential Evolution (BBO-DE) algorithm \cite{Boussaid2011}, and a recently introduced hybrid TLBO-Jaya algorithm \cite{Tsiflikiotis2017}.

However, to the best of the authors knowledge, this is the first time that LGSO problem-oriented algorithms are applied to the previously described optimal power allocation problem. The novelty in our work lies also on the fact that we address for the first time the optimal power allocation problem in higher dimensions.

In detail, we apply four state-of-art EAs, namely the memetic linear population size reduction and semi-parameter adaptation (MLSHADE-SPA) \cite{MLSHADE-SPA}, the contribution-based cooperative coevolution recursive differential grouping (CBCC-RDG3) \cite{CC-RDG3}, the differential grouping with spectral clustering-differential evolution cooperative coevolution (DGSC-DECC) \cite{DGSC}, and the enhanced adaptive differential evolution (EADE) \cite{EADE}.  We evaluate the above algorithms on the optimal power allocation problem in IoT networks with different settings and various dimensions.

\section{Problem Description}
We assume that a WSN, in which a decentralized detection scheme is utilized, has a fusion center and $L$ sensor nodes that are spatially distributed.
The observed signal vector can be expressed as $\mathbf{x}={{[{{x}_{1}},\,{{x}_{2}},...,{{x}_{L}}]}^{T}}$. Also, we assume that within the decentralized scheme two local observation states can occur, i.e. the absence or the presence of the signal. Thus, each node checks the binary hypothesis problem, by examining the hypotheses ${{H}_{0}}$ (absence of signal) and ${{H}_{1}}$ (presences of signal). The initial probabilities of the hypotheses are defined as $P\left( {{H}_{0}} \right)={{\pi }_{0}}$ and $P\left( {{H}_{1}} \right)={{\pi }_{1}}$.

If we consider the constant signal detection problem, in which an additive Gaussian noise is applied, and having the assumption that the local observation ${z}_{k}$ is achieved at the $k$-th node, the given problem can be formulated  as \cite{Boussaid2011}, \cite{Wimalajeewa2008}
\begin{eqnarray}
\begin{array}{l}
{H_0}\,:\,{{z}_{k} } = {{v}_{k} },\,\,\,k = 1,2,...,L\\
{H_1}\,:\,{{z}_{k} } = {{x}_{k} } + {{v}_{k} },\,\,\,k = 1,2,...,L
\end{array}
\end{eqnarray}	

Moreover, we assume that the additive local noise, ${{v}_{k}}$, is Gaussian distributed noise with mean value equal to zero and variance expressed as  $\sigma _{v}^{2}$. The signal ${{x}_{k}}$ is a known positive constant signal, so that ${{x}_{k}}=m,\,\,k=1,2,..,L$, for all sensors. The local observation of signal-to-noise ratio (SNR) is given by $\gamma ={{m}^{2}}/\sigma _{v}^{2}$. In a vector form, the observation of the SNR is expressed as
\begin{equation}
 \mathbf{z}=\mathbf{x}+\mathbf{v}
 \end{equation}
\noindent where $\mathbf{v}={{[{{v}_{1}},\,{{v}_{2}},...,{{v}_{L}}]}^{T}}$ is a Gaussian vector of a zero mean value and a covariance matrix ${\boldsymbol{\Sigma }}_{\mathbf{v}}$. Each sensor node applies a local decision ${{u}_{k}({z}_{k})}$. Additionally, an amplified version of the observation signal is re-transmitted to the fusion center by each node of the WSN \cite{Boussaid2011} and is given by
\begin{equation}
 {{u}_{k}({z}_{k})}={g}_{k}{z}_{k},\,\,k=1,2,..,L
 \end{equation}
\noindent where ${g}_{k}$ denotes the gain at the $k$-th node that is subject to attenuation and fading. Therefore, the received signal ${r}_{k}$, which is originated from the $k$-th node, arrives at the fusion center under the two previously mentioned hypotheses \cite{Boussaid2011, Wimalajeewa2008}. They are expressed as
\begin{eqnarray}\label{eq:two}
\begin{array}{l}
{H_0}\,:\,{{r}_{k}} = \,\,{{n}_{k} };\,\,\,k = 1,2,...,L\\
{H_1}\,:\,{{r}_{k} } = \,\,{{h}_{k}}{{g}_{k}}{{x}_{k}} + {{n}_{k}};\,\,\,k = 1,2,...,L
\end{array}
\end{eqnarray}
\noindent where
\begin{itemize}
	\item ${{h}_{k}}$ denotes the channel fading coefficient,
	\item ${{g}_{k}}$ denotes the gain,
	\item ${{n}_{k}}={{h}_{k }}{{g}_{k }}{{v}_{k}}+{{w}_{k}}$ is the effective noise at the fusion center having zero mean value, and
	\item ${{w}_{k}}$ is the receiver noise that can be analyzed to a sequence of independent and identically distributed components having a zero mean value and a variance equal to $\sigma _{w}^{2}$.
\end{itemize}
Therefore, the covariance matrix of the effective noise vector ${\mathbf{n}}$ can be expressed as
\begin{equation}\label{eq05}
{\boldsymbol{\Sigma }}_{\mathbf{n}}={{h}_{k }}{{g}_{k }}{\boldsymbol{\Sigma }}_{\mathbf{v}}{{g}_{k }}{{h}_{k }}+{\boldsymbol{\Sigma }}_{\mathbf{w}}
\end{equation}
 \noindent where ${\boldsymbol{\Sigma }}_{\mathbf{w}}={\sigma }_{\mathbf{w}}^{\mathbf{2}}\mathbf{I}$ is the covariance matrix of the receiver noise and $\mathbf{I}$ is the $L\times L$ identity matrix. In most of the cases, the observations that are recorded by the sensors are correlated. If we assume that between adjacent nodes the space is equally distributed in a straight line having a distance of $d$, the correlation between noise samples originated by nodes $i$ and $j$ is ${{\rho }^{d\left| i-j \right|}}$, with $\left| \rho \right|\le 1$.
In \ref{eq05}, ${{\boldsymbol{\Sigma }}_{\mathbf{v}}}$ can be expressed using a symmetric Hermitian Toeplitz matrix \cite{Boussaid2011, Wimalajeewa2008} in the following form
\begin{equation}\label{eq:four}
	{{\Sigma }_{v}}=\sigma _{v}^{2}\left[ \begin{matrix}
   1 & \rho^d  & .\,\,.\,\,. & {{\rho }^{d(L-2)}} & {{\rho }^{d(L-1)}}  \\
   \rho^d  & 1 & .\,\,.\,\,. & {{\rho }^{d(L-3)}} & {{\rho }^{d(L-2)}}  \\
   . & . & .\,\,.\,\,. & . & .  \\
   {{\rho}^{d(L-1)}} & {{\rho}^{d(L-2)}} & .\,\,.\,\,. & \rho^d & 1  \\
\end{matrix} \right] 	
\end{equation}
It is written in vector notation as
\begin{equation}
\mathbf{r=Ax+n}
 \end{equation}
\noindent where
\begin{itemize}
	\item $\mathbf{r}={{[{{r}_{1}},\,{{r}_{2}},...,{{r}_{L}}]}^{T}}$ denotes the received information vector,
	\item $\mathbf{n}={{[{{n}_{1}},\,{{n}_{2}},...,{{n}_{L}}]}^{T}}$ denotes the noise vector, and
	\item matrix $\mathbf{A}$ is given by $\mathbf{A}=diag\left( {{h}_{1}}{{g}_{1}},{{h}_{2}}{{g}_{2}},...,{{h}_{L}}{{g}_{L}} \right)$.
\end{itemize}

The observations $\textbf{r}$ at the fusion center are expressed as \cite{Boussaid2011, Wimalajeewa2008}
\begin{eqnarray}\label{eq:three}
 \begin{array}{l}
{H_0}:{\bf{r}}\sim N\left( {0,{{\boldsymbol{\Sigma }}_{\bf{n}}}} \right)\\
{H_1}:{\bf{r}}\sim N\left( {{\bf{Am,}}{{\boldsymbol{\Sigma }}_{\bf{n}}}} \right)
\end{array}
\end{eqnarray}
Moreover, the covariance matrix of the noise at the fusion center is given by \cite{Boussaid2011, Wimalajeewa2008}
\begin{equation}
{{\boldsymbol{\Sigma }}_{\mathbf{n}}}\mathbf{=A}{{\boldsymbol{\Sigma }}_{\mathbf{v}}}\boldsymbol{A+\sigma }_{\mathbf{w}}^{\mathbf{2}}\mathbf{I}
\end{equation}
\noindent where $\mathbf{m}$ is the $L$-length vector with all components equal to $m$. Taking into consideration the threshold $\tau ={{\pi }_{0}}/{{\pi }_{1}}$, and assuming the minimum probability of Bayesian fusion error, the optimum Bayesian decision rule is mathematically formulated as \cite{Wimalajeewa2008}
\begin{eqnarray}\label{eq:new}
\delta ({\bf{r}}) = \left\{ \begin{array}{l}
1\,\,\,\,if\,T({\bf{r}}) \ge \ln \tau \\
0\,\,\,if\,T({\bf{r}}) < \ln \tau
\end{array} \right.
\end{eqnarray}
The optimal procedure to apply the decision between the two hypotheses denotes the threshold rule on a log-likelihood ratio (LLR) of the observation vector. Thus, the LLR for the detection problem can be written as \cite{Boussaid2011}
\begin{equation}
T\left( {\bf{r}} \right) = m{{\bf{e}}^{\bf{T}}}{\boldsymbol f{A \Sigma }}_{\bf{n}}^{{\bf{ - 1}}}{\bf{r}} - \frac{1}{2}{m^2}{{\bf{e}}^{\bf{T}}}{\boldsymbol{A\Sigma }}_{\bf{n}}^{{\bf{ - 1}}}{\bf{Ae}}
\end{equation}
For the given detection problem, the LLR distribution under the two hypotheses is given by \cite{Boussaid2011, Wimalajeewa2008}
\begin{align}\label{eq:five}
{H_0}:T\left( {\bf{r}} \right)\sim N\left( { - \frac{1}{2}{m^2}{{\bf{e}}^{\bf{T}}}{\boldsymbol{A\Sigma }}_{\bf{n}}^{{\bf{ - 1}}}{\bf{Ae}},{m^2}{{\bf{e}}^{\bf{T}}}{\boldsymbol{A\Sigma }}_{\bf{n}}^{{\bf{ - 1}}}{\bf{Ae}}} \right)\\
{H_1}:T\left( {\bf{r}} \right)\sim N\left( {\frac{1}{2}{m^2}{{\bf{e}}^{\bf{T}}}{\boldsymbol{A\Sigma }}_{\bf{n}}^{{\bf{ - 1}}}{\bf{Ae}},{m^2}{{\bf{e}}^{\bf{T}}}{\boldsymbol{A\Sigma }}_{\bf{n}}^{{\bf{ - 1}}}{\bf{Ae}}} \right)
\end{align}
Based on \cite{Boussaid2011, Wimalajeewa2008, Tsiflikiotis2017}, if we further assume that the two hypotheses are equally probable, then the threshold is $\tau =1$. The fusion error probability, i.e. the fusion center selects ${{H}_{1}}$ when ${{H}_{0}}$ is true, or ${{H}_{0}}$ when ${{H}_{1}}$ is true, is expressed as
\begin{equation}\label{eq:Pe}
	{{P}_{e}}=Q\left( \frac{1}{2}\sqrt{{{m}^{2}}{{\mathbf{e}}^{\mathbf{T}}}\boldsymbol{A\Sigma }_{\mathbf{n}}^{\mathbf{-1}}\mathbf{Ae}} \right) 	
\end{equation}
\noindent where $Q\left( . \right)$ denotes the Gaussian Q-function. Therefore, the optimal power allocation problem is to obtain a set of $L$ optimal sensor gain values $\mathbf{g}=({{g}_{1}},{{g}_{2}},...,{{g}_{L}})$, which minimize the total power, by keeping the fusion error probability smaller than a specified criterion $\varepsilon $, \cite{Boussaid2011}. The mathematical formulation of the optimization problem can be expressed as
\begin{eqnarray}\label{eq:seven}
\begin{array}{l}
\min \,\,\,\,f\left( {\bf{g}} \right) = \sum\nolimits_{\ell = 1}^L {g_\ell ^2} \\
{}\,{}\,\,\,\,\zeta \left( {\bf{g}} \right) = Q\left( {\frac{1}{2}\sqrt {{m^2}{{\bf{e}}^{\bf{T}}}{\boldsymbol{A\Sigma }}_{\bf{n}}^{{\bf{ - 1}}}{\bf{Ae}}} } \right) - \varepsilon \le 0\\
\,\,\,\,\,\,\,\,\,\,\,\,\,\,\,\,\,\,\,\,\,\,\,\,{\psi _l}\left( {\bf{g}} \right) = - {g_\ell } \le 0,\,\,l = 1,2,...,L
\end{array}
\end{eqnarray}

If we apply a penalty function to the above optimization problem, we can combine the objective and constraint functions to a single objective function. As in \cite{Tsiflikiotis2017}, we utilize a dynamically modified penalty function, because of its effectiveness against a static penalty function approach \cite{Parsopoulos2002}. As a result, the objective function is formulated as
\begin{eqnarray}\label{eq:obj}
\begin{array}{l}
F\left( {\bf{g}} \right) = f\left( {\bf{g}} \right) + It \left[ {\sum\limits_{i = 0}^L {\theta \left( {{q_i}\left( {\bf{g}} \right)} \right) {q_i}{{\left( {\bf{g}} \right)}^{\lambda \left( {{q_i}\left( {\bf{g}} \right)} \right)}}} } \right]\\
{q_i}\left( {\bf{g}} \right) = \left\{ \begin{array}{l}
\max \left\{ {0,\zeta \left( {\bf{g}} \right)} \right\},\,\,\,\mathrm{if}\,i = 0\\
\max \left\{ {0,{\psi _i}\left( {\bf{g}} \right)} \right\},\,\,\mathrm{otherwise}
\end{array} \right.\\
\lambda \left( x \right) = \left\{ \begin{array}{l}
1\,\,\,\,if\,x < 1\\
2\,\,\,otherwise
\end{array} \right.\\
\theta \left( x \right) = \left\{ \begin{array}{l}
10\,\,\,\,\,if\,x \le 0.1\\
100\,\,\,\,\,\,if\,\,\,0.1 < x \le 1\\
300\,\,\,otherwise\,
\end{array} \right.
\end{array}
\end{eqnarray}
\noindent where
\begin{itemize}
	\item $It$ denotes the current iteration number of the algorithm,
	\item $\theta \left( {{q}_{i}}\left( \mathbf{g} \right) \right)$ is a multi-stage assignment function, and
	\item $\lambda \left( {{q}_{i}}\left( \mathbf{g} \right) \right)$ represents the power of the penalty function.
\end{itemize}
\begin{table*}[htbp]
  \centering
  \caption{Algorithms average results for $L=300$ sensors, $\gamma=10dB$. The smaller values are in bold font.}
    \begin{tabular}{c c c c c c}
    \hline
    \multicolumn{6}{c}{\textbf{D=300}} \bigstrut\\
    \hline
    \textbf{$\rho $} & \multicolumn{1}{l}{\textbf{$\varepsilon$}} & \multicolumn{1}{l}{\textbf{MLSHADE-SPA}} & \multicolumn{1}{l}{\textbf{DGSC-DECC}} & \multicolumn{1}{l}{\textbf{CBCC-RDG3}} & \multicolumn{1}{l}{\textbf{EADE}} \bigstrut
    \\ \hline
    \multirow{4}{*}{0}          & 0.1           & \textbf{1.785} & 208.517   & 19.600          & 680.660 \\
                            & 0.05          & 3.078          & 213.802   & \textbf{1.754}  & 734.090 \\
                            & 0.01          & 7.966          & 238.207   & \textbf{4.246}  & 713.802 \\
                            & 0.001         & 15.750         & 206.634   & \textbf{8.862}  & 718.025 \\
\multirow{4}{*}{0.01}       & 0.1           & \textbf{1.709} & 200.217   & 19.583          & 752.375 \\
                            & 0.05          & 3.240          & 208.556   & \textbf{1.789}  & 741.533 \\
                            & 0.01          & 7.950          & 221.228   & \textbf{4.260}  & 761.676 \\
                            & 0.001         & 14.745         & 212.173   & \textbf{8.922}  & 697.972 \\
\multirow{4}{*}{0.1}        & 0.1           & \textbf{2.105} & 186.664   & 19.576          & 718.065 \\
                            & 0.05          & 3.436          & 215.164   & \textbf{1.818}  & 719.622 \\
                            & 0.01          & 8.157          & 206.904   & \textbf{4.454}  & 689.980 \\
                            & 0.001         & 16.081         & 243.721   & \textbf{9.395}  & 729.020 \\
\multirow{4}{*}{0.5}        & 0.1           & \textbf{2.372} & 195.501   & 21.634          & 726.920 \\
                            & 0.05          & 3.568          & 242.939   & \textbf{2.125}  & 862.677 \\
                            & 0.01          & 8.267          & 228.090   & \textbf{5.595}  & 729.278 \\
                            & 0.001         & 17.143         & 212.832   & \textbf{12.701} & 702.204 \\ \hline
    \end{tabular}%
  \label{tab:300}%
\end{table*}%
\begin{table*}[htbp]
  \centering
  \caption{Algorithms average results for $L=600,800$ sensors, $\rho=0$, $\gamma=10dB$. The smaller values are in bold font.}
    \begin{tabular}{c c c c c c}
    \hline
    \textbf{$D $} & \multicolumn{1}{l}{\textbf{$\varepsilon$}} & \multicolumn{1}{l}{\textbf{MLSHADE-SPA}} & \multicolumn{1}{l}{\textbf{DGSC-DECC}} & \multicolumn{1}{l}{\textbf{CBCC-RDG3}} & \multicolumn{1}{l}{\textbf{EADE}} \bigstrut
    \\ \hline
    \multirow{4}{*}{600} & 0.1   & \textbf{2.727} & 11937.294 & 573.896        & 12344.849 \\
                     & 0.05  & 4.789          & 11964.444 & \textbf{1.802} & 12447.193 \\
                     & 0.01  & 9.971          & 11944.247 & \textbf{3.838} & 12420.642 \\
                     & 0.001 & 18.762         & 12012.850 & \textbf{7.530} & 12385.260 \\
\multirow{4}{*}{800} & 0.1   & \textbf{3.623} & 7789.121  & 45000.000      & 9943.722  \\
                     & 0.05  & 5.432          & 8088.043  & \textbf{2.189} & 10126.987 \\
                     & 0.01  & 11.164         & 8043.257  & \textbf{4.594} & 10845.345 \\
                     & 0.001 & 20.813         & 7399.785  & \textbf{8.789} & 10041.129 \\\hline
    \end{tabular}%
  \label{tab:600}%
\end{table*}%
\begin{table}[htbp]
  \centering
  \caption{Average Rankings achieved by Friedman test.}
    \begin{tabular}{l c c c}
     \hline
    \textbf{Algorithm} & \textbf{Average Rank} & \textbf{Normalized values} & \textbf{Rank} \\ \hline
MLSHADE-SPA        & 1.75                  & 1.31                       & 2             \\
DGSC-DECC          & 2.96                  & 2.22                       & 3             \\
CBCC-RDG3          & 1.33                  & 1.00                       & 1             \\
EADE               & 3.96                  & 2.97                       & 4    \\    \hline
    \end{tabular}%
  \label{tab:Friedman}%
\end{table}%
\begin{table}[htbp]
  \centering
  \caption{Wilcoxon signed-rank test between CBCC-RDG3 and the other algorithms.The bold font indicates values below significance level.}
    \begin{tabular}{lc}
     \hline
\textbf{CBCC-RDG3   vs} & \textbf{p-value} \\  \hline
MLSHADE-SPA             & \textbf{3.426E-02}        \\
DGSC-DECC               & \textbf{4.000E-05 }       \\
EADE                    & \textbf{7.000E-06}          \\    \hline
    \end{tabular}%
  \label{tab:pval}%
\end{table}%
\begin{figure}[!htbp]
\centering
\subfloat[]{\includegraphics[width=\columnwidth]{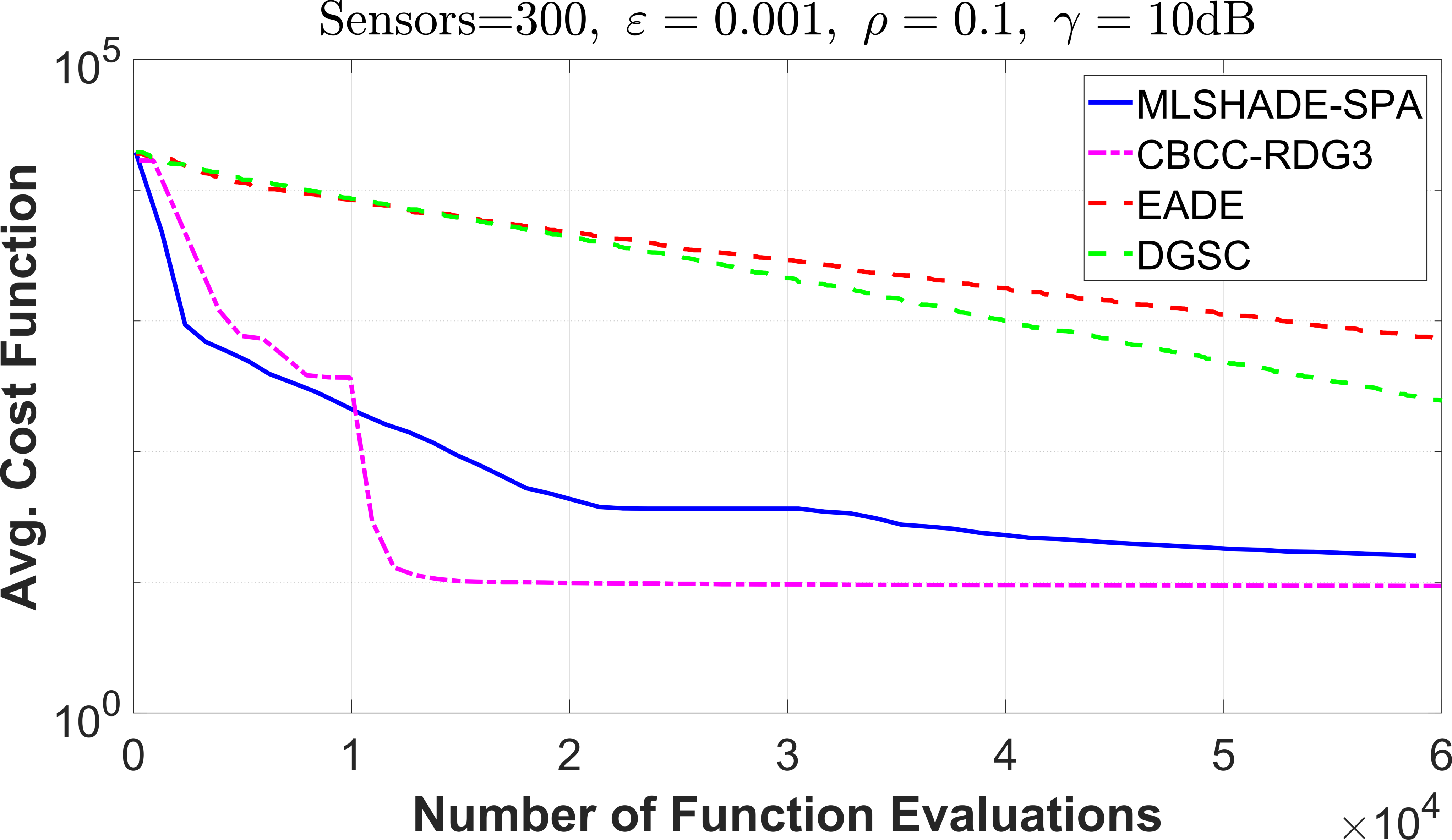}%
 \label{Convplot300_1}}
\hfil
\subfloat[]{\includegraphics[width=\columnwidth]{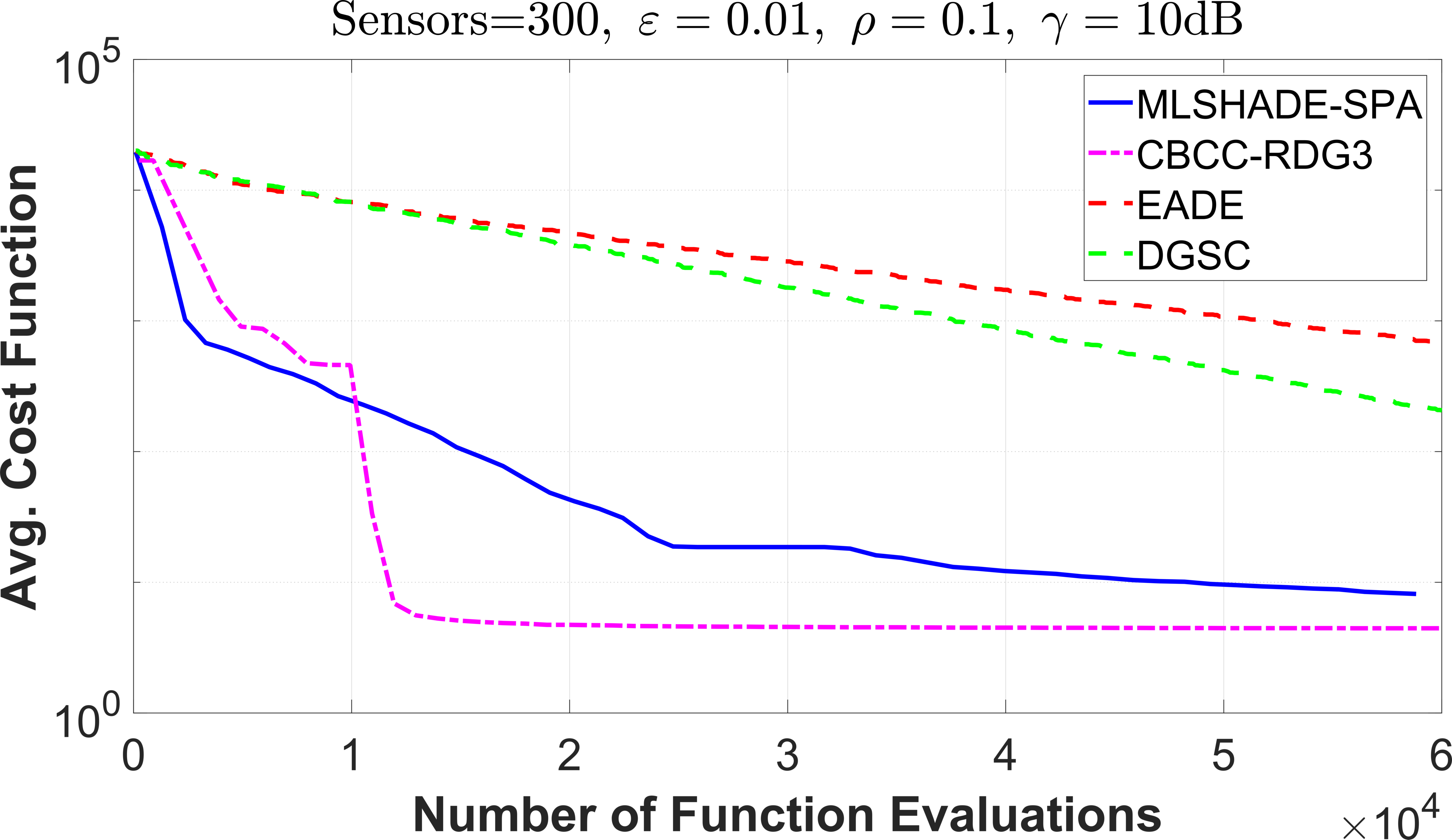}%
 \label{Convplot300_2}}
\hfil
\subfloat[]{\includegraphics[width=\columnwidth]{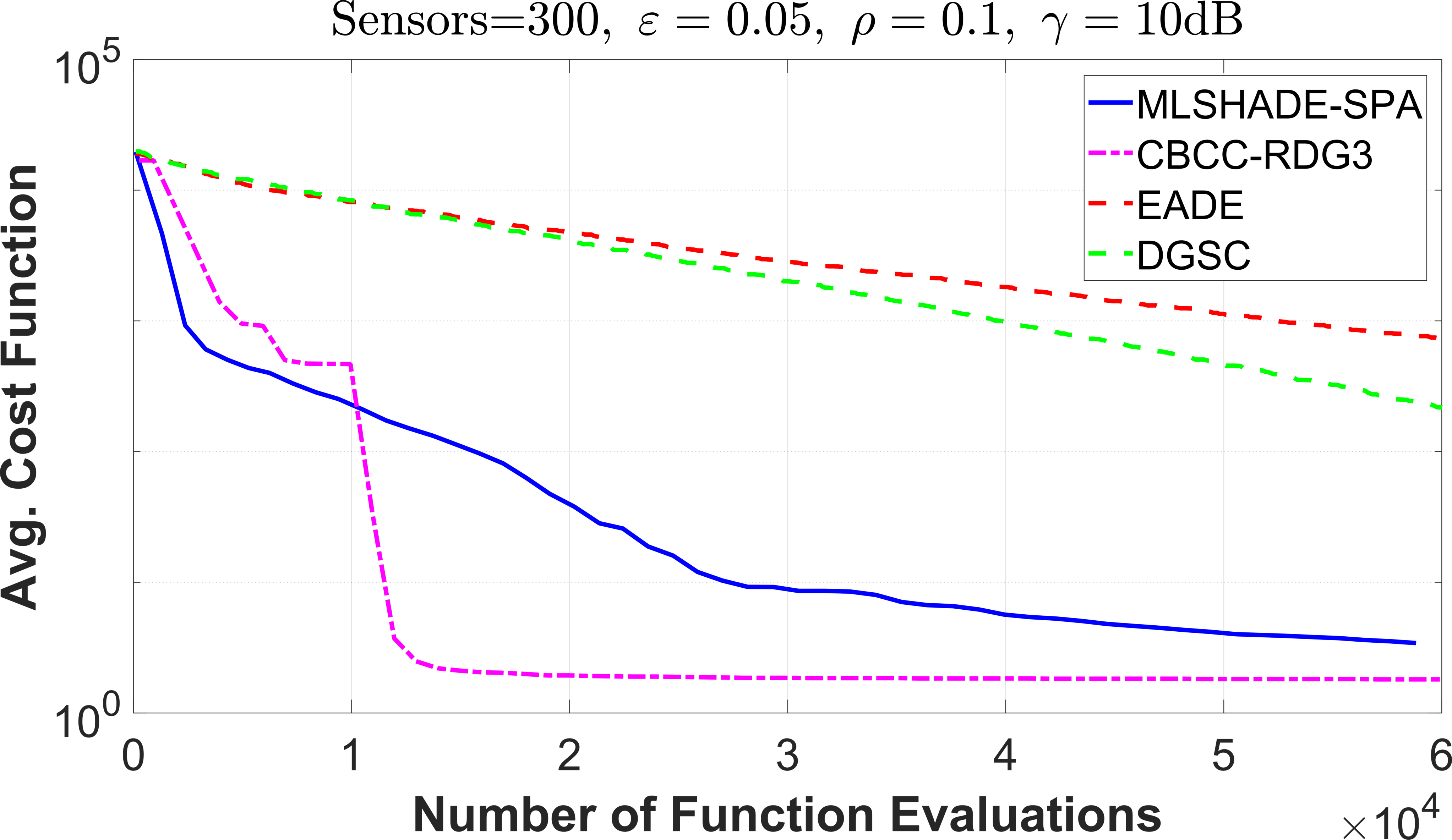}%
\label{Convplot300_3}}
\hfil
\subfloat[]{\includegraphics[width=\columnwidth]{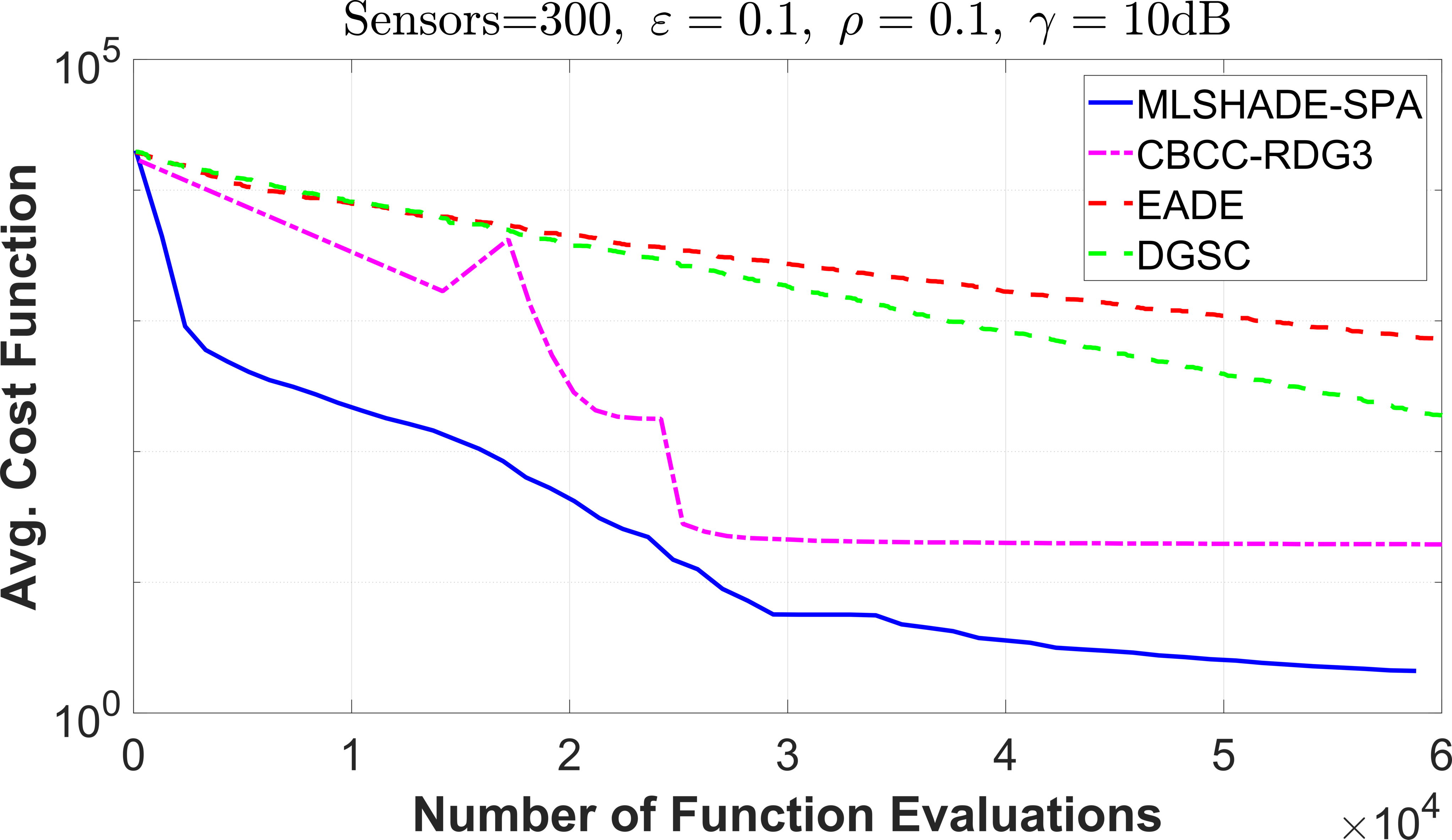}%
\label{Convplot300_3}}
\caption{Convergence rate graph for $L=300, \gamma =10\,\mathrm{dB}, \rho=0.1, a)\varepsilon=0.001, b)\varepsilon=0.01, c)\varepsilon=0.05, d)\varepsilon=0.1$.}
\label{fig1}
\end{figure}
\begin{figure}[!htbp]
\centering
\subfloat[]{\includegraphics[width=\columnwidth]{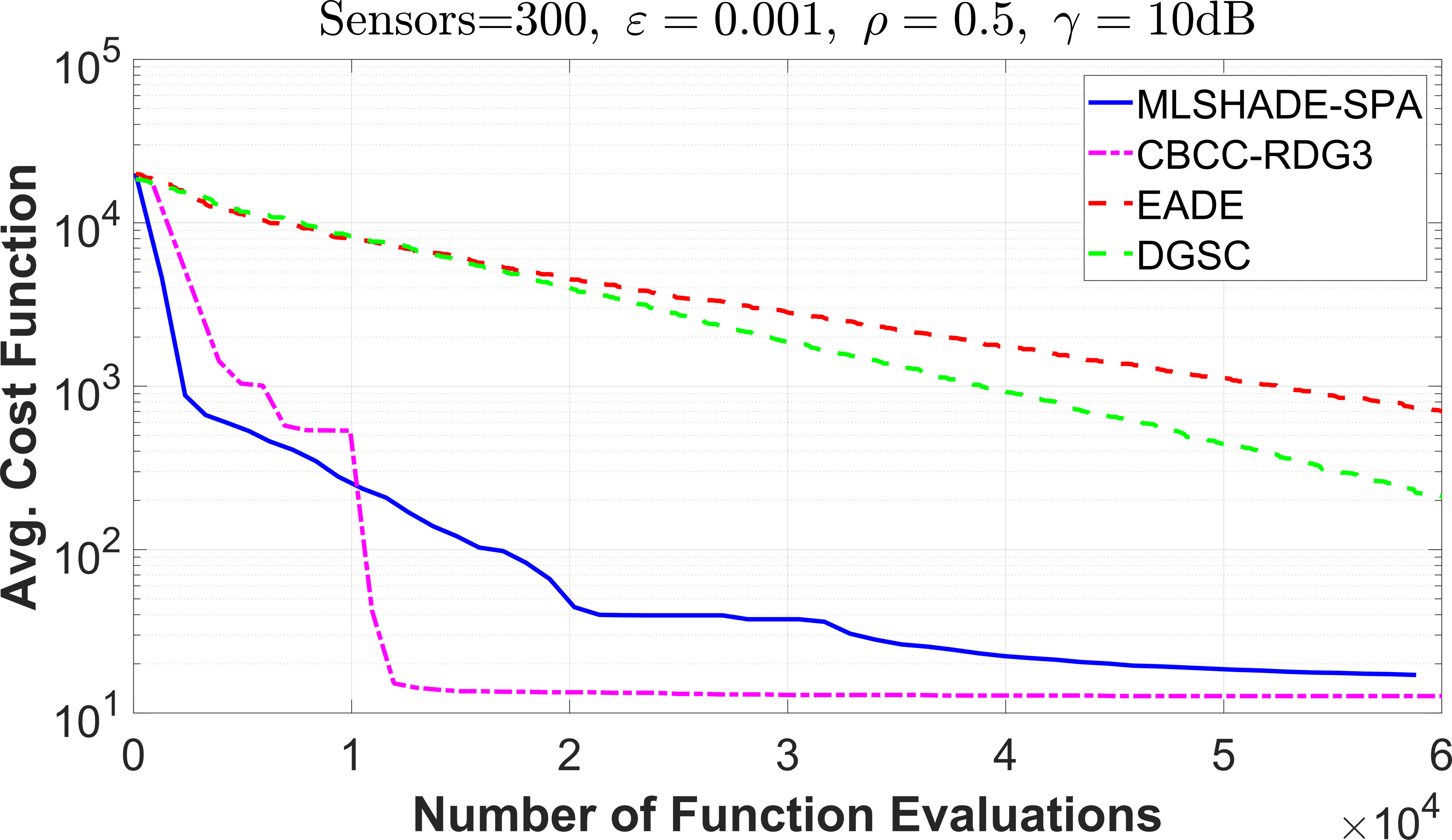}%
 \label{Convplot300_5}}
\hfil
\subfloat[]{\includegraphics[width=\columnwidth]{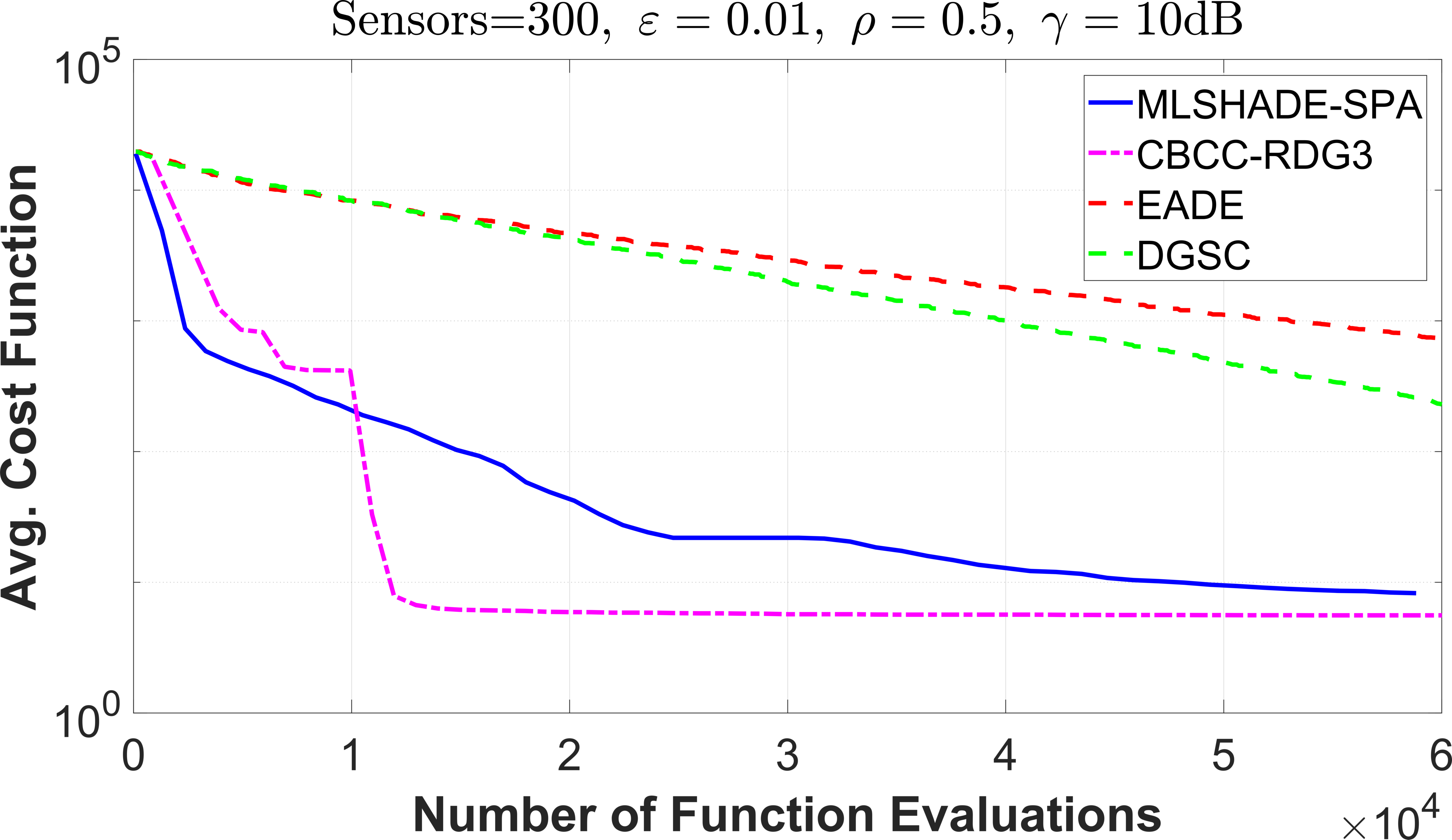}%
 \label{Convplot300_6}}
\hfil
\subfloat[]{\includegraphics[width=\columnwidth]{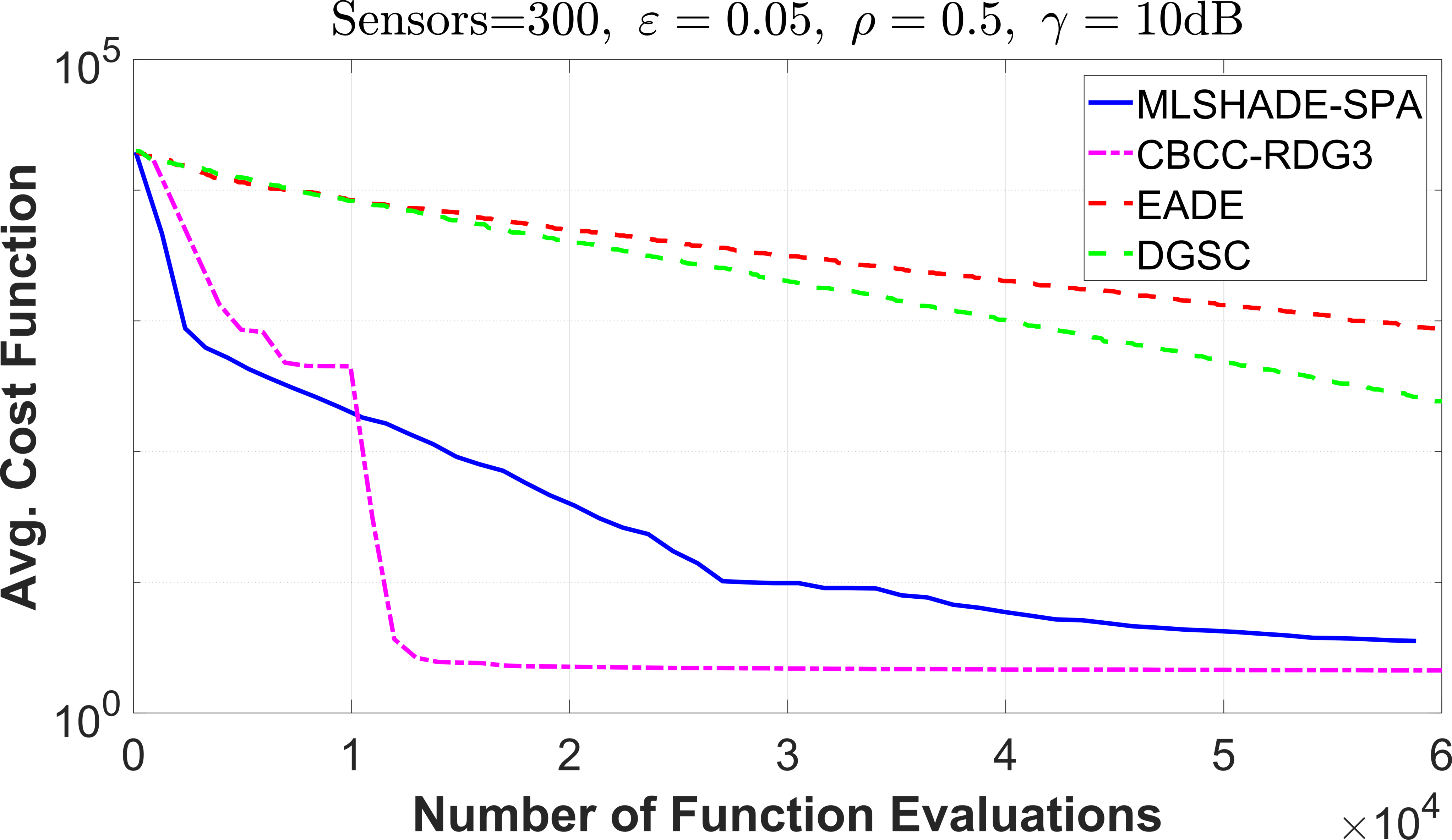}%
\label{Convplot300_7}}
\hfil
\subfloat[]{\includegraphics[width=\columnwidth]{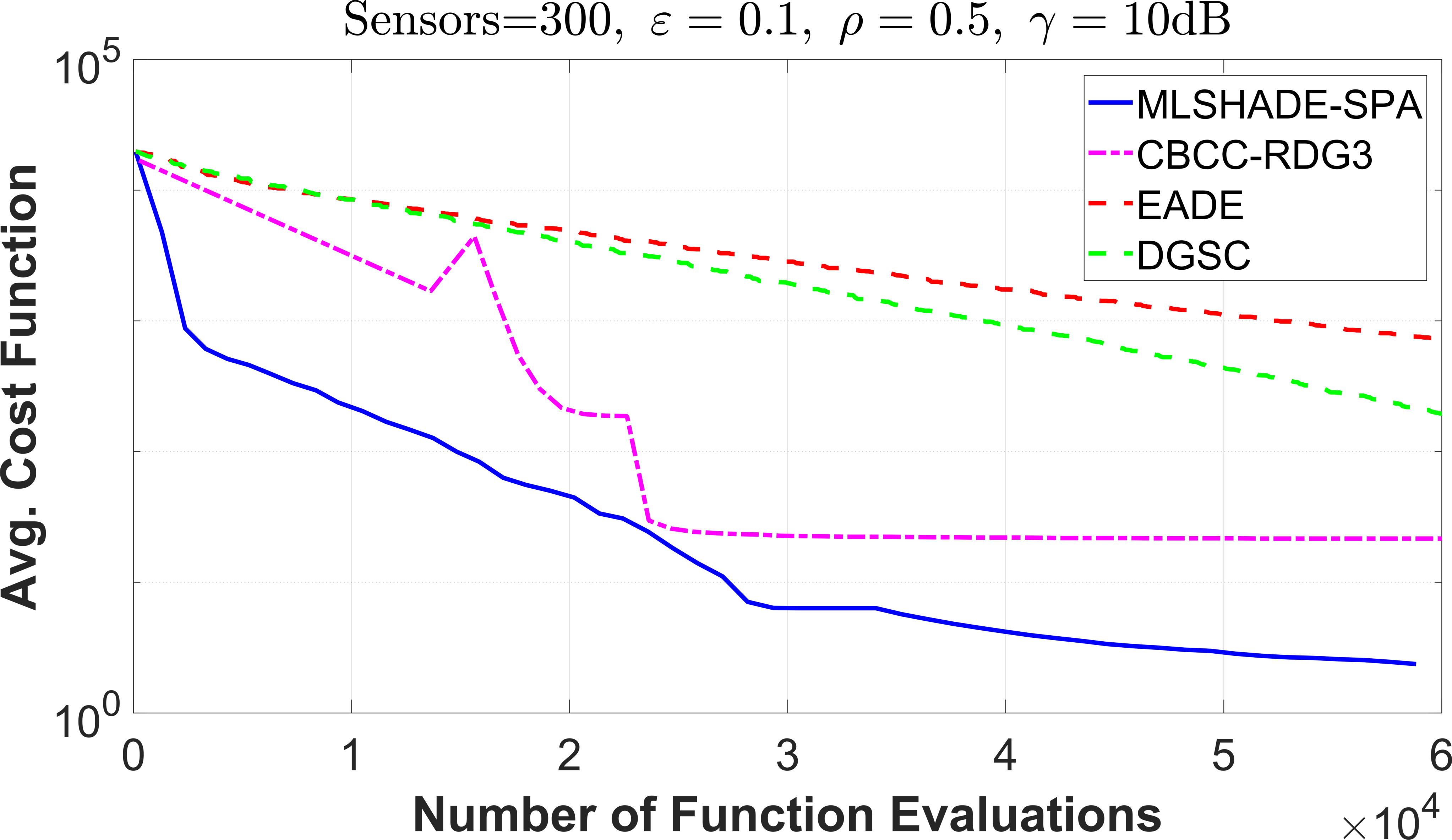}%
\label{Convplot300_8}}
\caption{Convergence rate graph for $L=300, \gamma =10\,\mathrm{dB}, \rho=0.5, a)\varepsilon=0.001, b)\varepsilon=0.01, c)\varepsilon=0.05, d)\varepsilon=0.1$.}
\label{fig2}
\end{figure}
\begin{figure}[!htbp]
\centering
\subfloat[]{\includegraphics[width=\columnwidth]{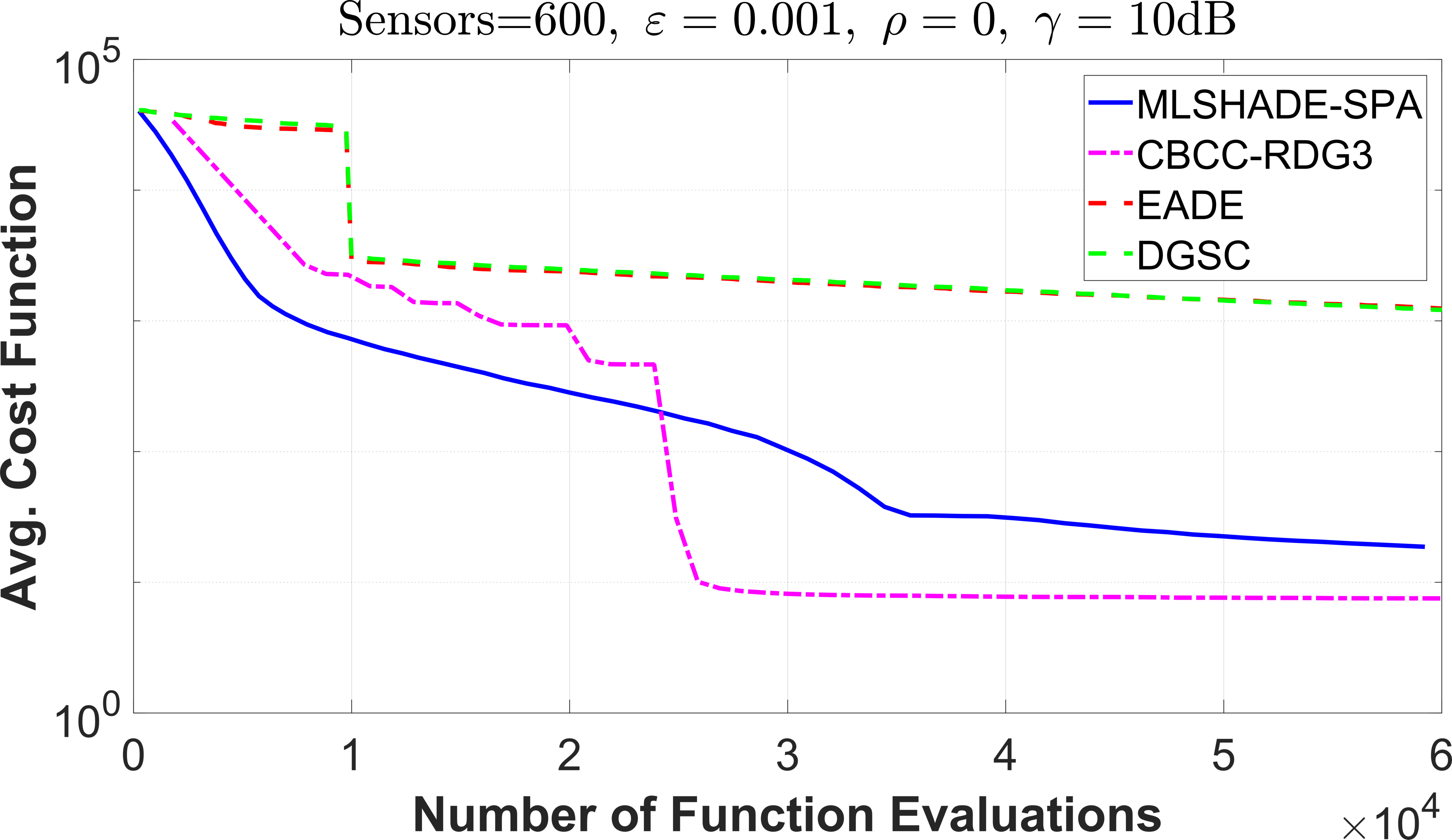}%
 \label{Convplot600_1}}
\hfil
\subfloat[]{\includegraphics[width=\columnwidth]{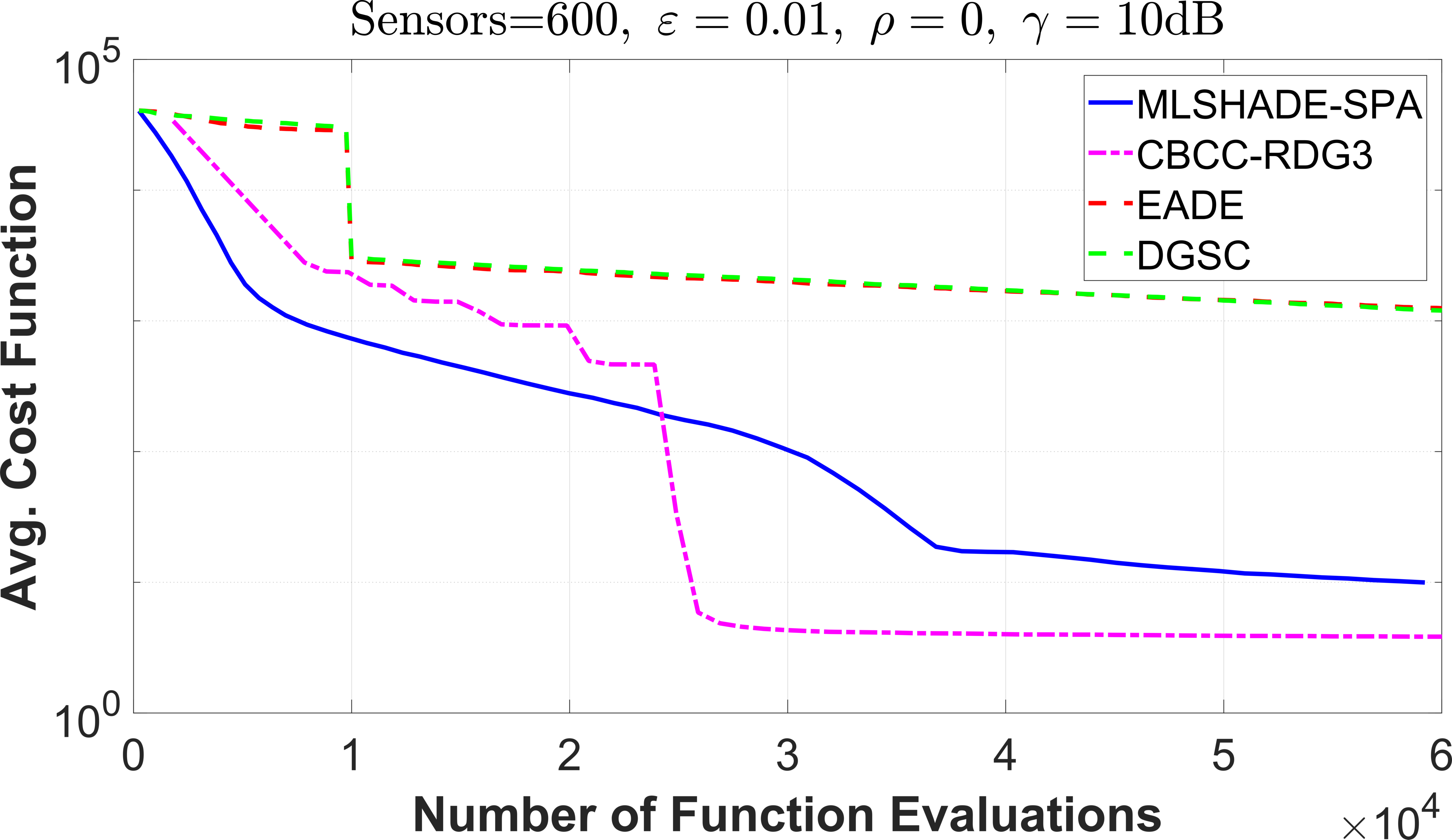}%
 \label{Convplot600_2}}
\hfil
\subfloat[]{\includegraphics[width=\columnwidth]{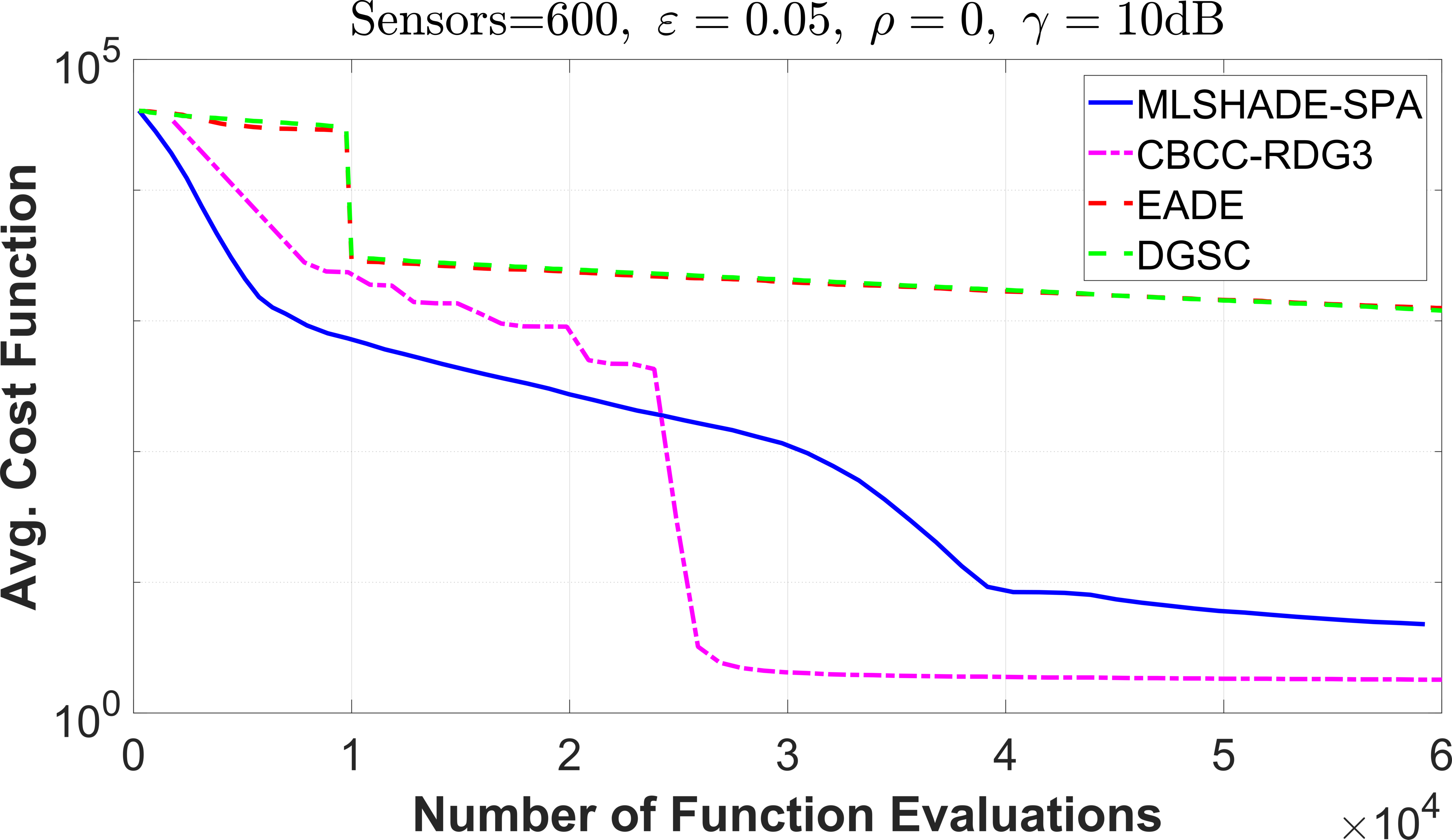}%
\label{Convplot600_3}}
\hfil
\subfloat[]{\includegraphics[width=\columnwidth]{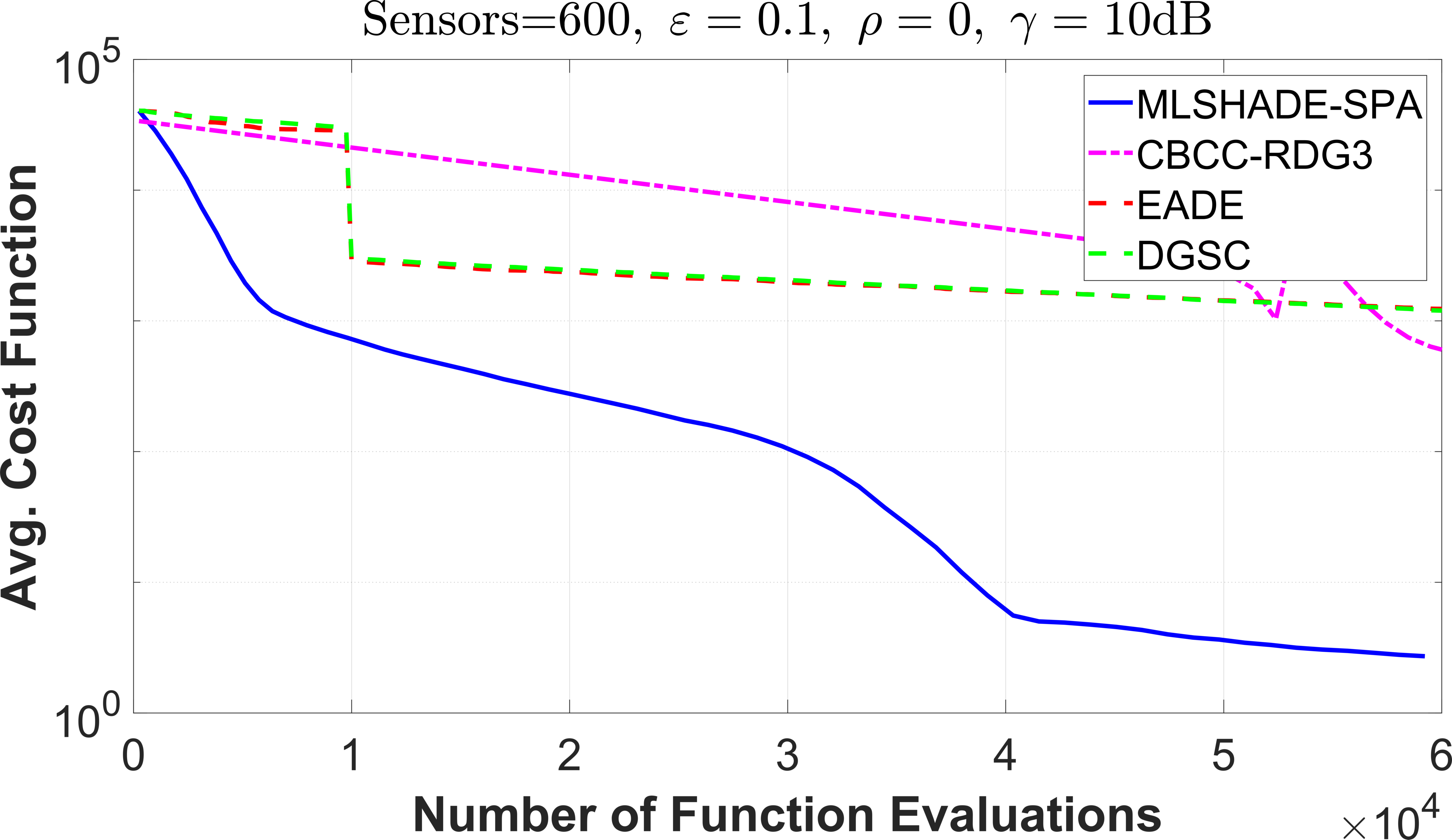}%
\label{Convplot600_4}}
\caption{Convergence rate graph for $L=600, \gamma =10\,\mathrm{dB}, \rho=0, a)\varepsilon=0.001, b)\varepsilon=0.01, c)\varepsilon=0.05, d)\varepsilon=0.1$.}
\label{fig3}
\end{figure}
\begin{figure}[!htbp]
\centering
\subfloat[]{\includegraphics[width=\columnwidth]{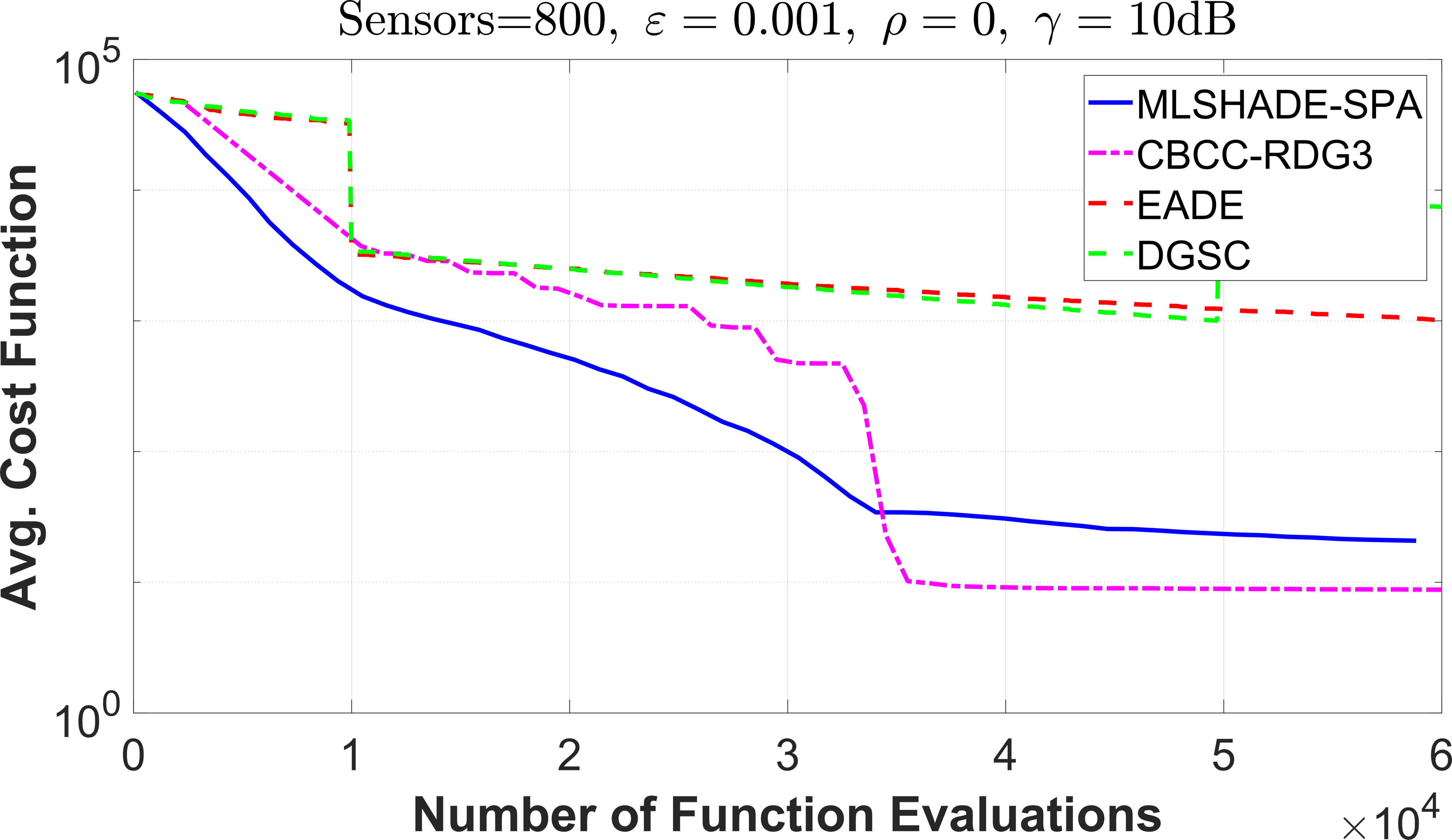}%
 \label{Convplot800_1}}
\hfil
\subfloat[]{\includegraphics[width=\columnwidth]{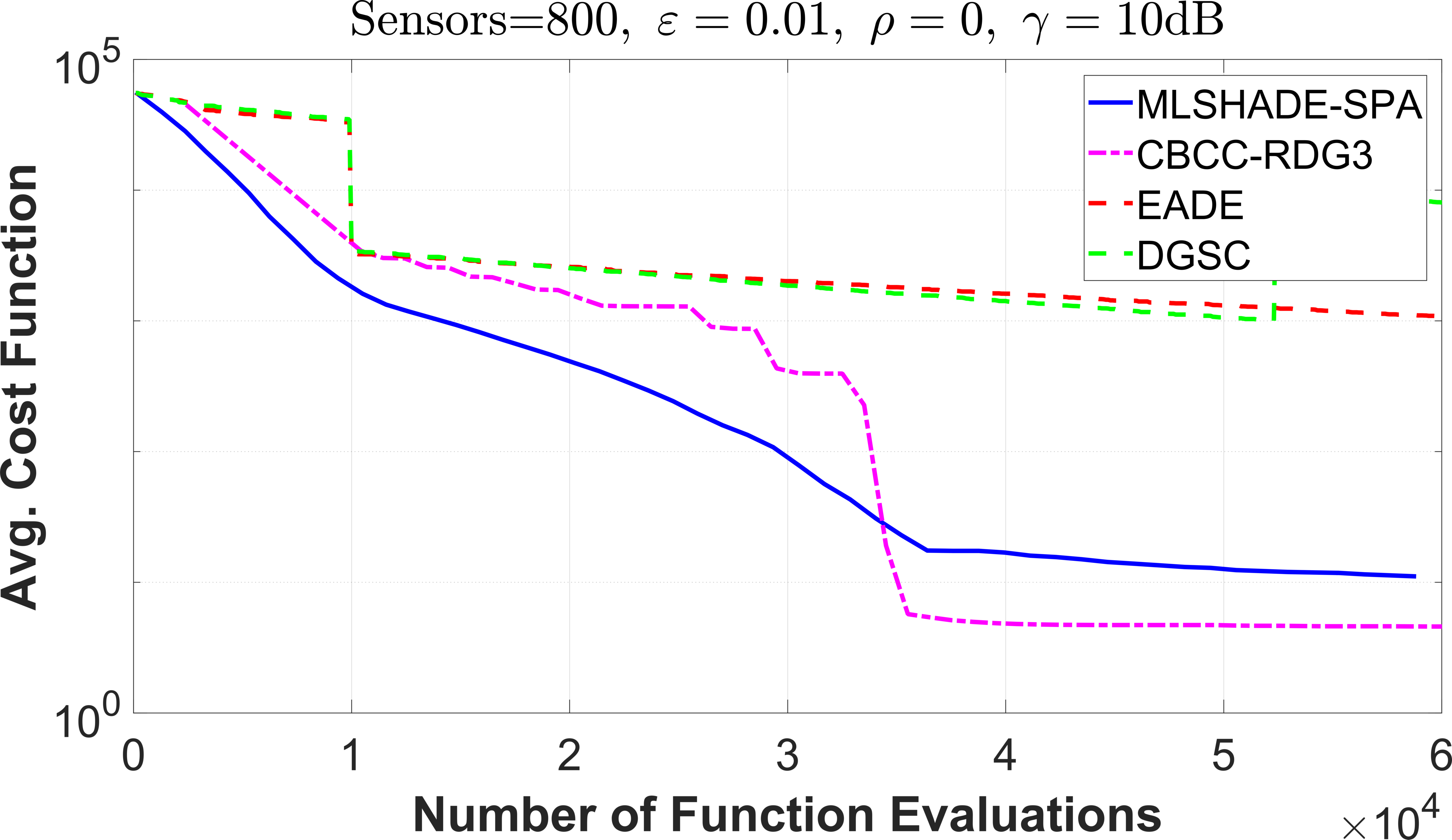}%
 \label{Convplot800_2}}
\hfil
\subfloat[]{\includegraphics[width=\columnwidth]{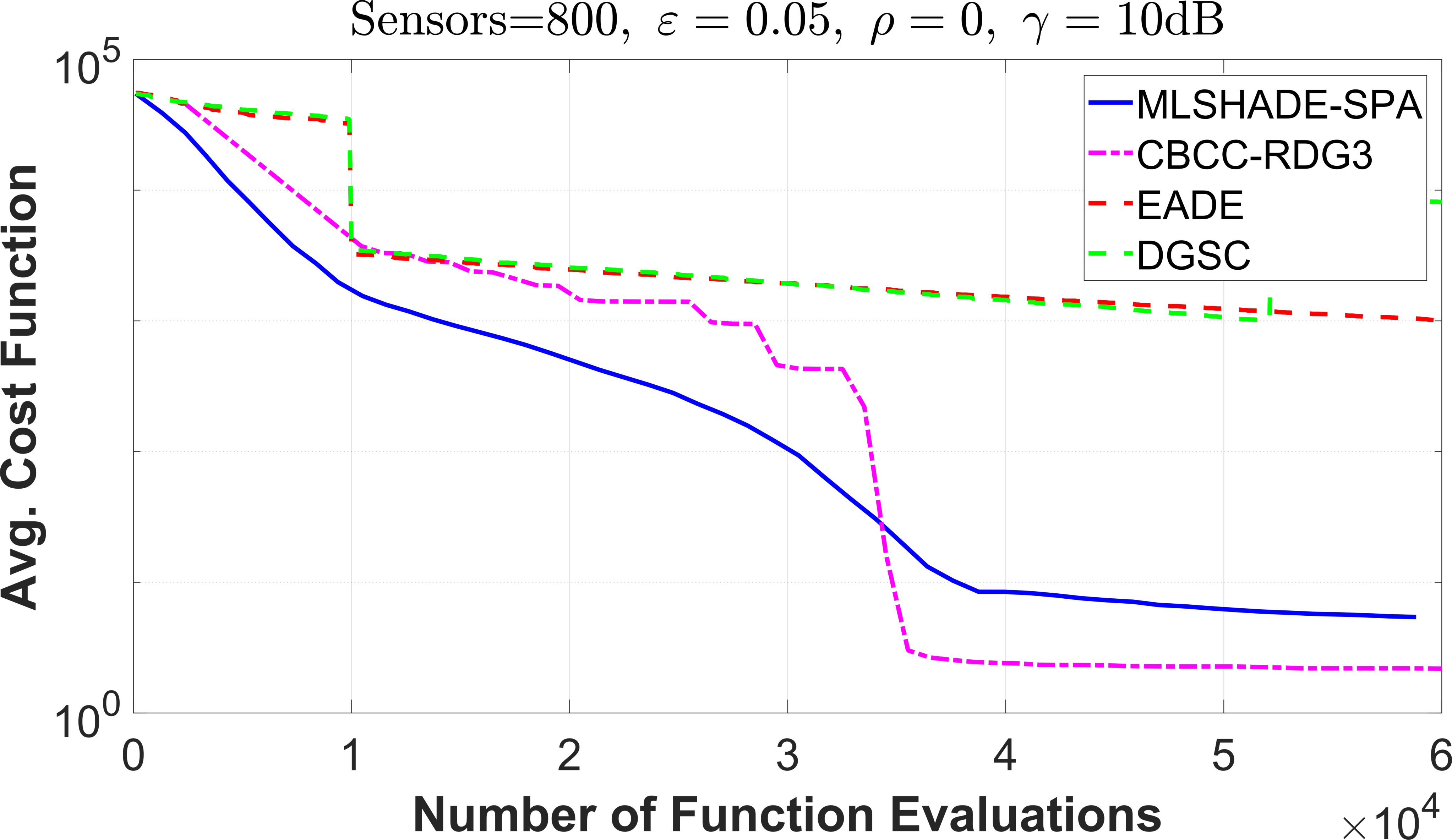}%
\label{Convplot800_3}}
\hfil
\subfloat[]{\includegraphics[width=\columnwidth]{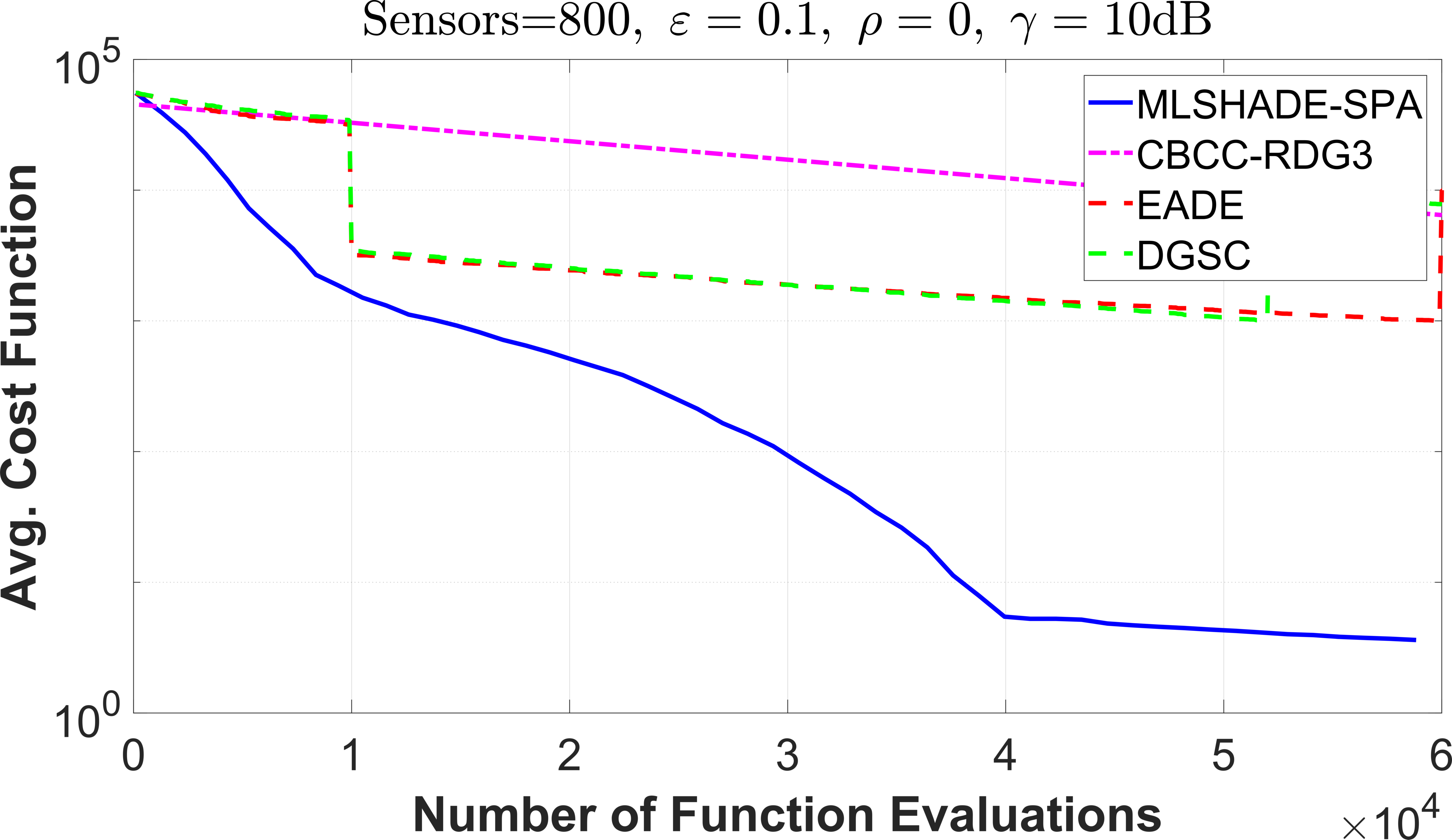}%
\label{Convplot800_4}}
\caption{Convergence rate graph for $L=800, \gamma =10\,\mathrm{dB}, \rho=0, a)\varepsilon=0.001, b)\varepsilon=0.01, c)\varepsilon=0.05, d)\varepsilon=0.1$.}
\label{fig4}
\end{figure}

\section{Large Scale Global Optimization algorithms}
\label{algorithm_description}
The algorithms that we will apply for the optimal power allocation problem can be classified into two categories; those that solve the problem directly, either by using several algorithms and their hybridization, and those that use a decomposition strategy, as well as a variable grouping strategy. The first category includes the MLSHADE-SPA \cite{MLSHADE-SPA} and EADE \cite{EADE} algorithms, whereas CBCC-RDG3 \cite{CC-RDG3} and DGSC-DECC \cite{DGSC} algorithms fall into the second category.

\subsection{Enhanced adaptive differential evolution}
Mohamed \cite{EADE} proposed an enhanced adaptive differential evolution (EADE) algorithm, which is a non-Cooperative Co-evolution method, to provide feasible solutions in large scale optimization problems. EADE includes two novel contributions. Firstly, in order to benefit from the available information of the whole population, a novel mutation scheme has been introduced. The proposed scheme utilizes two vectors, which are randomly selected, of the top and bottom 100p\% individuals in the current population of size $NP$. The third vector is randomly selected from the middle [NP-2(100p\%)] individuals. A combination between the mutation rule and the basic mutation strategy \emph{DE/rand/1/bin} is applied. It is worth noting that the probability between the mutation rules is equal ($0.5$). Secondly, a novel self-adaptive scheme has been introduced that utilizes the gradual change of the crossover rate values. This scheme, during the evolution process, can benefit from the previous experience of the individuals in the search space. This feature, on its turn, can balance effectively the trade-off between the convergence speed and the population diversity.

\subsection{MLSHADE-SPA}
Hadi et al. \cite{MLSHADE-SPA} proposed an LSHADE-SPA memetic framework, which was the runner up in CEC2018 Large Scale global optimization competition. MLSHADE-SPA exhibits hybrid characteristics by combining population-based algorithms and local search. LSHADE-SPA \cite{LSHADE-SPA}, EADE \cite{EADE}, and ANDE \cite{ANDE} are utilized as population-based algorithms for global exploration, whereas a modified version of MTS (MMTS) is utilized as a local search algorithm for local exploitation. In order to further enhance the framework's performance, the 'divide and conquer' concept is applied. This concept is processed without any prior assumptions of the optimized problems' structure; the dimensions of the problems are divided into groups in a random way and each of groups is solved independently.

\subsection{CBCC-RDG3}
A 'divide-and-conquer' approach is proposed in \cite{CC-RDG3} to address large-scale optimization problems having components with overlapping behavior. The main idea behind the 'divide-and-conquer' approach is to decompose overlapping problems by modifying the Recursive Differential Grouping (RDG) method. In order to do so, the linkage to shared by multiple components variables must be broken. The previously mentioned procedure of decomposition by utilizing RDG method can be described using the following three steps. The first step is to distinguish the interaction between the decision variables and the selected variable $z_i$, in order to classify them into a subset $Z_1$. The second step distinguishes and groups the interaction between the decision variables and any other variable in $Z_1$ recursively, in order $Z_1$ to becoming independent from the remaining variables. The third step repeats the two previous steps until all variables have been grouped. As a result, in the case of an overlapping problem, an assignment into a single group for all the decision variables will occur, since all the variables will be linked.

RDG3, which is designed for overlapping problems, modifies the above procedure by further examining the interaction between the set $Z_1$ and the rest of the available variables (excluding $Z_1$) to identify those that interact with $z_i$ indirectly. If the algorithm finds any interaction, the interacting decision variables are moved to $Z_1$. RDG3 repeats these steps until there is no detection  of interaction between $Z_1$ and the rest of the available variables. The decision variables in $Z_1$ are considered to be a consolidated group.

The RDG3 method moves on to the next decision variable $z_i$ that has not been distinguished and grouped. The above process is repeated until all the decision variables have been classified in groups. RDG3 continues to further divide the separable variables into small groups with an interval $e_s$. Upon completion of RDG3 method, the identified separable and consolidated variable groups are returned. The contribution-based CC (CBCC) basic idea is to allocate available computational resources to components, based on how they contribute to the overall fitness improvement. This is grouped with RDG3 to form the CBCC-RDG3 algorithm \cite{CC-RDG3}. The selected solver is the covariance matrix adaptation- evolutionary strategy (CMA-ES) \cite{CMA-ES}, which utilizes the round-robin scheme to optimize sub-problems. The computational resources to each sun-problem, by utilizing the round-robin scheme, are equally distributed. CBCC-RDG3 was the winner in IEEE CEC 2019 LSGO competition.

\subsection{DGSC-DECC algorithm}
The authors in \cite{DGSC} present a differential grouping with spectral clustering (DGSC) algorithm. The basic concept of Differential Grouping (DG) is to find the interaction relationship between the available variables according to their differential values. On the other hand, Spectral Clustering (SC) is graph partition problem, which is based on spectral graph theory \cite{SC}, to provide optimal solutions. Spectral clustering forms an undirected weighted graph, by considering all the data as edge-connected points in space. The weight of the edge originates from the similarity value between the two points. Based on these concepts, the authors in \cite{DGSC} assume that the decision variables are points in space and consider the interaction relationship between these variables as the weight of the edge. DGSC combines differential grouping with spectral clustering, where the similarity matrix of spectral clustering comes from the differential values of grouping. The final result is the grouping of the problem's decision variables. The sub-component optimizer that is utilized, is described in \cite{SaNSDE}. This resulted algorithm is called DGSC-DECC. DGSC-DECC was the runner-up in IEEE CEC 2019 LSGO competition.

\section{Numerical Results}
In this section, the numerical results of the algorithms are presented.
We apply four LSGO algorithms to different WSN configurations, the MLSHADE-SPA \cite{MLSHADE-SPA}, the EADE \cite{EADE}, the CBCC-RDG3 \cite{CC-RDG3}, and the DGSC-DECC \cite{DGSC}. For all of the selected algorithms, we set the population size equal to $100$ and the maximum number of objective function evaluation to $60,000$. All the selected algorithms are executed for $50$ independent trials. During the initialization of each algorithm, the population is randomly selected, based on the lower and upper boundaries. The number of objective-function evaluations is applied as the stopping criterion to the optimization problem. The initial position (solution in the optimization process) of each member of the population in each dimension is selected within the range [0, 15].

We apply the objective function (cost function) defined in \eqref{eq:obj}.
Additionally, and without losing the generality of the problem, we introduce the assumption that the coefficients ${{h}_{i}}$, which describe the channel fading, have a Rayleigh distribution with a unit mean value. Moreover, we also assume that they are ranked in a descending order ${{h}_{1}}\ge {{h}_{2}}\ge ...\ge {{h}_{L}}$.

The corresponding simulations of the first case are performed by applying the following parameters
\begin{itemize}
	\item Number of sensor nodes $L=\{300\}$,
	\item Different fusion error threshold values \\
     $\varepsilon=\{0.001,0.05,0.01,0.1\}$, and
	\item Correlation factor values $\rho=\{0,0.01,0.1,0.5\}$.
\end{itemize}
Both values of different fusion error threshold and correlation factor are applied for each of the sensor nodes. The $SNR$ is set to a fixed value of $10$dB. Considering all the above, we obtain $4$ different optimization cases for each $\rho$ value. As a result, the number of dimensions of the total problem results in 300.

Table~\ref{tab:300} lists the average results of all executions of the algorithms. We notice that CBCC-RDG3 outperforms the other algorithms in 12 out of the 16 cases. MLSHADE-SPA performs better in four cases having $\varepsilon=0.1$. However, MLSHADE-SPA obtains a solution in all cases. The other algorithms fail to produce acceptable solutions and the objective function values are significantly high.

In the second case, the following parameters are modified
\begin{itemize}
	\item Number of sensor nodes $L=\{600,800\}$,
	\item Different fusion error threshold values \\ $\varepsilon=\{0.001,0.05,0.01,0.1\}$, and
	\item Correlation factor $\rho=0$ (Uncorrelated case).
\end{itemize}
In this case, we set the population size to $250$, whereas the maximum number of objective function evaluations remains $60,000$. Taking into consideration the given parameters, the optimization problem becomes more difficult for the selected algorithms to solve. The comparative results of the algorithms are listed in Table~\ref{tab:600}. Once again, we notice that CBCC-RDG3 performs better in six out of the eight cases, however it fails to obtain an acceptable solution in two cases. MLSHADE-SPA manages to obtain a feasible solution in all cases. EADE and DGSC-DECC fail to obtain acceptable solutions and achieve higher objective function values. In this case, the algorithms' results seem to be worse than the previous one.

A comparison between the algorithms in terms of two non-parametric statistical tests is also performed. The rankings according to Friedman test are reported in Table~\ref{tab:Friedman}. It is noteworthy that the CBCC-RDG3 obtained the best rank of $1.33$, having the MLSHADE-SPA the second best rank of $1.75$. The Wilcoxon signed-rank test between the CBCC-RDG3 and the other selected algorithms is listed in Table~\ref{tab:pval}. The $p$-values below 0.05, which denoted the level of significance, are indicated in boldface. Again, we can observe that the CBCC-RDG3 for the optimal power allocation optimization problem is performed significantly better than all the other algorithms.

The convergence rate plots for $L=\{300\}$ and $\rho=\{0.1,0.5\} $ are depicted in Figs.~\ref{fig1} and \ref{fig2}, respectively. We notice that the CBCC-RDG3 converges faster than the other algorithms in all of the given cases except two. MLSHADE-SPA converges with a similar speed in all cases and achieves a small value of the objective function. Both algorithms converge at similar speed at higher objective function values. Moreover, Figs.~\ref{fig3} and \ref{fig4} portray the corresponding convergence rate graphs for $L=\{600, 800\}$ and $\rho=0$. We can observe that the plots are similar with the previous case. The only difference is that the CBCC-RDG3 fails to obtain an acceptable solution in two cases, which is evident by Figs. ~\ref{Convplot600_4} and \ref{Convplot800_4}.

Concluding our results, it can be clearly seen that MLSHADE-SPA is more stable than CBCC-RDG3 algorithm with the different fusion error threshold $\varepsilon=\{0.001,0.05,0.01,0.1\}$, as it reaches very promising solutions in all cases. Besides, having a fusion error threshold $\varepsilon=\{0.1\}$ in all dimensions, MLSHADE-SPA outperforms CBCC-RDG3 algorithm, which proves that MLSHADE-SPA can reach the global solution in stochastic environment, i.e. it is able to solve very complicated stochastic real-world problems.

\section{Conclusion}
In this paper, we have applied LSGO algorithms to the optimal power allocation problem in IoT networks. We have compared results with a small number of objective function evaluations and different numbers of sensors. Overall, CBCC-RDG3 is a powerful optimizer that obtains good results in most of the cases. However, MLSHADE-SPA obtains a solution in every case, which could be very useful in a real-case of wireless communications, where decisions in real-time are often desirable. EADE and DGSC-DECC failed to obtain acceptable solutions, which indicates that more objective function evaluations are probably required. DGSC-DECC performance can be explained since this algorithm is designed to perform well with overlapping variables, which is not the case here. In our future work, we will apply LSGO algorithms to more demanding optimization problems in wireless communications.

\section*{Acknowledgment}
This project has received funding from the European Union’s Horizon 2020 research and innovation programme under grant agreement No. 957406 (TERMINET).
%
%
%
%

\bibliographystyle{IEEEtran}
%

\bibliography{CEC2021}

\end{document}